\documentclass[pdflatex,sn-nature]{sn-jnl1}

\usepackage{graphicx}
\usepackage{multirow}
\usepackage{amsmath,amssymb,amsfonts}
\usepackage{amsthm}
\usepackage{mathrsfs}
\usepackage[title]{appendix}
\usepackage{textcomp}
\usepackage{manyfoot}
\usepackage{booktabs}
\usepackage{xcolor}
\usepackage{algorithm}
\usepackage{algorithmicx}
\usepackage{algpseudocode}
\usepackage{listings}
\usepackage{verbatim}
\usepackage{tabularray}
\usepackage{xcolor}%

\theoremstyle{thmstyleone}

\theoremstyle{thmstyletwo}%
\theoremstyle{thmstylethree}%
\NewDocumentCommand{\todo}
{ mO{} }{\textcolor{magenta}{\textsuperscript{\textit{TODO}}\textsf{\textbf{\small[#1]}}}}

\begin{document}

\title[Assessing and Understanding Creativity in Large Language Models]{Assessing and Understanding Creativity in Large Language Models}

\author[1]{\normalsize \fnm{Yunpu} \sur{Zhao}}

\author[2]{\normalsize \fnm{Rui} \sur{Zhang}}
\author[3,5]{\normalsize \fnm{Wenyi} \sur{Li}}

\author[2]{\normalsize \fnm{Di} \sur{Huang}}

\author[2]{\normalsize \fnm{Jiaming} \sur{Guo}}

\author[3]{\normalsize \fnm{Shaohui} \sur{Peng}}

\author[2]{\normalsize \fnm{Yifan} \sur{Hao}}

\author[2]{\normalsize \fnm{Yuanbo} \sur{Wen}}

\author[2,4]{\normalsize \fnm{Xing} \sur{Hu}}

\author[2,4]{\normalsize \fnm{Zidong} \sur{Du}}

\author[2]{\normalsize \fnm{Qi} \sur{Guo}}

\author[3,5]{\normalsize \fnm{Ling} \sur{Li}}

\author*[2,5]{\normalsize \fnm{Yunji} \sur{Chen}}\email{cyj@ict.ac.cn}

\affil[1]{\orgdiv{\normalsize University of Science and Technology of China},  \orgaddress{\city{Hefei}, \country{China}}}

\affil[2]{\orgdiv{\normalsize State Key Lab of Processors, Institute of Computing Technology}, \orgname{Chinese Academy of Sciences}, \orgaddress{\city{Beijing}, \country{China}}}

\affil[3]{\orgdiv{\normalsize Institute of Software}, \orgname{Chinese Academy of Sciences}, \orgaddress{\city{Beijing}, \country{China}}}

\affil[4]{\orgdiv{\normalsize Shanghai Innovation Center for Processor Technologies}, \orgaddress{\city{Shanghai}, \country{China}}}

\affil[5]{\orgdiv{\normalsize University of Chinese Academy of Sciences}, \orgaddress{\city{Beijing}, \country{China}}}


\abstract{}
In the field of natural language processing, the rapid development of large language model (LLM) has attracted more and more attention. 
LLMs have shown a high level of creativity in various tasks, but the methods for assessing such creativity are inadequate. The assessment of LLM creativity needs to consider differences from humans, requiring multi-dimensional measurement while balancing accuracy and efficiency.
This paper aims to establish an efficient framework for assessing the level of creativity in LLMs.
By adapting the modified Torrance Tests of Creative Thinking, the research evaluates the creative performance of various LLMs across 7 tasks, emphasizing 4 criteria including Fluency, Flexibility, Originality, and Elaboration. In this context, we develop a comprehensive dataset of 700 questions for testing and an LLM-based evaluation method.
In addition, this study presents a novel analysis of LLMs' responses to diverse prompts and role-play situations.
We found that the creativity of LLMs primarily falls short in originality, while excelling in elaboration. Besides, the use of prompts and the role-play settings of the model significantly influence creativity. Additionally, the experimental results also indicate that collaboration among multiple LLMs can enhance originality.
Notably, our findings reveal a consensus between human evaluations and LLMs regarding the personality traits that influence creativity.
The findings underscore the significant impact of LLM design on creativity and bridges artificial intelligence and human creativity, offering insights into LLMs' creativity and potential applications.

\maketitle
\section{Introduction}\label{sec1}
In recent years, the realm of artificial intelligence (AI) has witnessed a meteoric rise in the development and sophistication of Large Language Models (LLMs)\cite{gpt4,llama2}. 
LLMs have significantly advanced in their capabilities in addressing a variety of conventional natural language processing tasks, such as reasoning and natural language understanding \cite{Reasoning,NLU,bert,NLG2}. 
Moreover, LLMs have also demonstrated significant value in widespread applications. 
From transforming rudimentary text into compelling narratives\cite{poetry,narration}, unlocking a new realm of storytelling, to solving complex algorithmic problems\cite{Code}, these models have shown a semblance of what could be interpreted as creativity. 
The practical manifestations of this creativity have penetrated various sectors, including science research, where they assist in idea generation and suggestion\cite{idea}; education, by providing personalized learning experiences\cite{education}; and in the entertainment industry, creating music and art\cite{dance,music}. 
In many of their applications, LLMs seem to exhibit the ability to generate original text, aiding tasks related to imagination and creativity, suggesting that they may indeed possess elements of creativity.

From the broad capabilities demonstrated by LLMs, the creativity they exhibit is a key reason they are considered powerful. However, behind the impressive abilities of LLMs lies a significant question that warrants careful examination: do these models actually possess real creativity, or is their apparent smartness merely an illusion—a complex imitation of human thinking created by their training paradigm? This question touches on the very nature of LLM intelligence, which may not be easily explained. Since LLMs have shown considerable creativity, understanding the extent and characteristics of this creativity is essential. Gaining deeper insight into the creativity of LLMs can not only guide us in further improving their performance but also in enhancing our understanding of the nature of their creativity. This, in turn, informs our daily use and application of these models, underscoring the need for an effective method to measure and assess their creativity.

Creativity, as a term, traditionally refers to the natural ability to think innovatively, to make unconventional connections, and to devise solutions that are both novel and effective\cite{creadef}.
Assessing the creativity of LLMs is fraught with challenges.
Firstly, the question of creativity does not have clear answers to refer to.
When we ask a LLM a question like ``What is the speed of light in vacuum in meters per second'', the answer can be formally vetted, given the objective nature of the topic. However, when posed with a prompt such as ``What would be the implications if animals could talk?'', the situation becomes different in this case because there is no definitive answer and the answer is open and divergent, making it challenging to judge the correctness of the output\cite{NMICorre}. 
Additionally, since creativity encompasses various aspects, including originality and flexibility, it is necessary to design diverse tasks and criteria to effectively measure these qualities in LLMs. 
Besides, there are differences between LLMs and humans, which might lead to irrelevant responses or serious logical issues, requiring us to additionally assess these aspects.
Finally, evaluating creativity necessitates a delicate balance between accuracy and efficiency, rendering traditional human-based evaluation methods less practical. 
Therefore, it is imperative to address the challenges outlined above to make a robust and sound assessment of creativity in LLMs.

Recognizing the need for a comprehensive assessment of LLM's creativity, we design an efficient framework to automatic assess the creativity of LLMs by adapting and modifying the Torrance Tests of Creative Thinking (TTCT)\cite{TTCT}, a widely recognized tool  in psychometrics' research for human creativity assessment. 
To enhance the credibility of the results and reduce the randomness, seven verbal tasks, which use verbal stimuli, were selected. 
We employed GPT-4, the most advanced LLM, to expand the question set for each task, thereby constructing the testing dataset.
To ensure a thorough and objective evaluation of creativity and capture creativity's various manifestations, we combine diver tasks and criteria. We design a comprehensive test protocol incorporating four criteria for measuring creativity: Fluency, Flexibility, Originality, and Elaboration.
We let the LLMs answer questions from the constructed dataset, obtaining many question-answer pairs.
we utilized GPT-4 as an evaluator to assess each answer, as GPT-4 is capable of effectively assessing the openness of responses and identifying their shortcomings and errors. Under proper prompt engineering, GPT-4 can efficiently and effectively complete the evaluation of the entire results of the dataset. Thus, we can achieved a balance between efficiency and accuracy in our assessment method. 

We selected six popular LLMs as test subjects, each possessing different architectures and parameter scales. In addition to the overall testing, we conducted some additional exploratory experiments that investigating the changes of creativity levels exhibits by LLMs when given different types of prompts and different roles that LLMs played. Then, we designed a collaboration mechanism for LLMs to explore the impact of multiple LLMs collaborating on creativity. Lastly, we also performed some psychological experiments related to personality traits on the LLMs, including emotional intelligence (EI), empathy, the Big Five Inventory (BFI) and self-efficacy. Because we found in relevant psychological research that human creativity is correlated with these personality traits and we verified the consistency between LLMs and humans in this regard.

Our experiments and analysis have yielded several conclusions. Firstly, there are significant differences in the creative performance among different models, even among those of the same scale with an equal number of parameters. This variation primarily exists between different type of models. Their differences are mainly reflected in the model architecture, parameter settings during training, alignment strategies, and the datasets used for training. 
Additionally, we observed that models generally excel in the Elaboration metric, but tend to be less adept in demonstrating Originality. 
Besides, The type of prompt and the specific role-play request given to the model also play a significant role in influencing its creative output. When the models are given instructive prompts or Chain-of-Thought prompts, there is a significant increase in the level of creativity. Also, having LLM play different roles leads to notable differences; the role of a scientist demonstrates the highest level of creativity. Many roles even show a decrease compared to the default scenario, but there is generally an improvement in originality. Then, collaboration among multiple LLMs can enhance the level of creativity, with the most notable improvement in originality.
Lastly, the psychological scale results reveal a consistency between LLMs and humans in terms of associated creativity factors, such as emotional intelligence (EI), empathy, self-efficacy, and others.

\section{Human creativity: assessment and related personality traits}\label{sec2}
\subsection{Creativity assessment in psychological research}\label{sec21}
The question of creativity assessment has been a prominent focus on the creativity research, especially since the 1950s, marking the inception of a systematic study into individual differences in creativity\cite{creass}. 
For example, Guilford pioneered the research on creativity and his famous structure of intellect model was mainly about defining and analyzing the factors constitute intelligence, where creativity plays a major driving force in his theory\cite{guilford}.
In recent years, many new developments regarding the measurement of divergent thinking, consensual assessment technique and subjective ratings, and self-report methodology \cite{DT1,DT2,DT3}.
Although advances in methodology and technology have led to important developments regarding creativity assessment, some assessment methods have long been described as "gold standard" for creativity assessment\cite{Cress,apadebate}.
Among them, TTCT\cite{TTCT} has been the most widely used and researched test of creativity, having extensive data to support its reliability and validity. Research about TTCT report good reliability scores for scoring and test-retest reliability\cite{validity}.

TTCT is designed to identify and assess an individual's creative potential by exploring various dimensions. Contrasting conventional assessments that emphasize convergent thinking, the test foster divergent thinking, encouraging participants to generate multiple solutions to open-ended, ambiguous problems.
TTCT has been widely applied in educational settings, organizational assessments, demonstrating its versatility and comprehensive approach to measuring creativity. Its ability to tap into various facets of creative thinking has made TTCT a reliable and respected tool\cite{TTCTReview}.
Due to the authority and comprehensiveness of the TTCT, we select tasks from the TTCT to construct our dataset.

\subsection{Creativity and personality: findings in psychological research}\label{sec22}
Research has revealed that creativity is not solely a fixed human personality trait. It evolves from a combination of individual processes such as cognitive, affective, behavioral, and contextual factors. Some psychologists have conducted a detailed meta-analysis of papers exploring the relationship between creativity and various personality traits\cite{meta1,meta2}. 

These studies' results highlight a correlation between creativity and a plethora of personal factors. Notably, elements such as emotional intelligence, divergent thinking, openness to experience, and intrinsic motivation stand out as strong influencers. However, factors like age, intelligence, and gender exhibits a relatively milder association with creativity, signifying a varied spectrum of influence across different personal traits.
Since large language models have exhibited some personality traits, we conducted experiments to test whether these findings also hold true in LLMs.

\subsection{Assessing the creativity of large language models}\label{sec23}
The emergence of abilities from LLMs continually surpass people's expectations, and the evaluation of various abilities of LLMs has received widespread attention\cite{evalurvey}. Currently, most evaluations focus on the ability of LLMs to solve tasks, with fewer evaluations combining aspects of psychology.

Although there are some studies focused on the intersection of LLM with psychology and cognitive science\cite{psychologyAI}, work discussing the creativity of LLM is still in a relatively early stage. Current studies are somewhat focused on exploring the creativity of LLMs, primarily from the standpoint of creativity theory, which aims to elucidate the definitions and challenges of applying creativity theory within the contexts of LLMs\cite{crea4}. Some initial evaluations of creativity in LLMs have also been undertaken\cite{crea1,crea2,crea3}. But these works only employed simple tasks such as Alternative Uses Task (AUT) to assess their creativity and lack the comparison between various LLMs leads to the limitation of the conclusions. It is worth mentioning that in \cite{TTCTTest}, the authors used the standard TTCT to assess GPT-4's creativity. The results showcased that GPT-4 achieved human top 1\% levels in fluency and originality, along with a high score in flexibility. This study leans more towards comparing advanced Large Language Models (LLMs) with human benchmarks. The original TTCT test protocol does not seamlessly adapt to assessing creativity in LLMs, as the limited sample of questions could induce randomness and accidental outcomes, making hypothesis testing challenging when comparing different models. Furthermore, expanding the number of question sets leads to exorbitant time costs in human-based evaluations.

Due to the differences between the human and LLM, it is problematic to directly use the TTCT's test protocol to benchmark LLMs' creativity. To address the dilemma, we proposed a new framework to systematically analyze LLM's creativity. This framework comprises carefully crafted metrics used in TTCT and a dataset that accounts for seven tasks. We will dive into details of the framework in next section.

\section{Overview of the framework}\label{sec3}
\begin{figure}[htbp]
    \centering
        \makebox[\textwidth][c]{\includegraphics[width=1.1\textwidth]{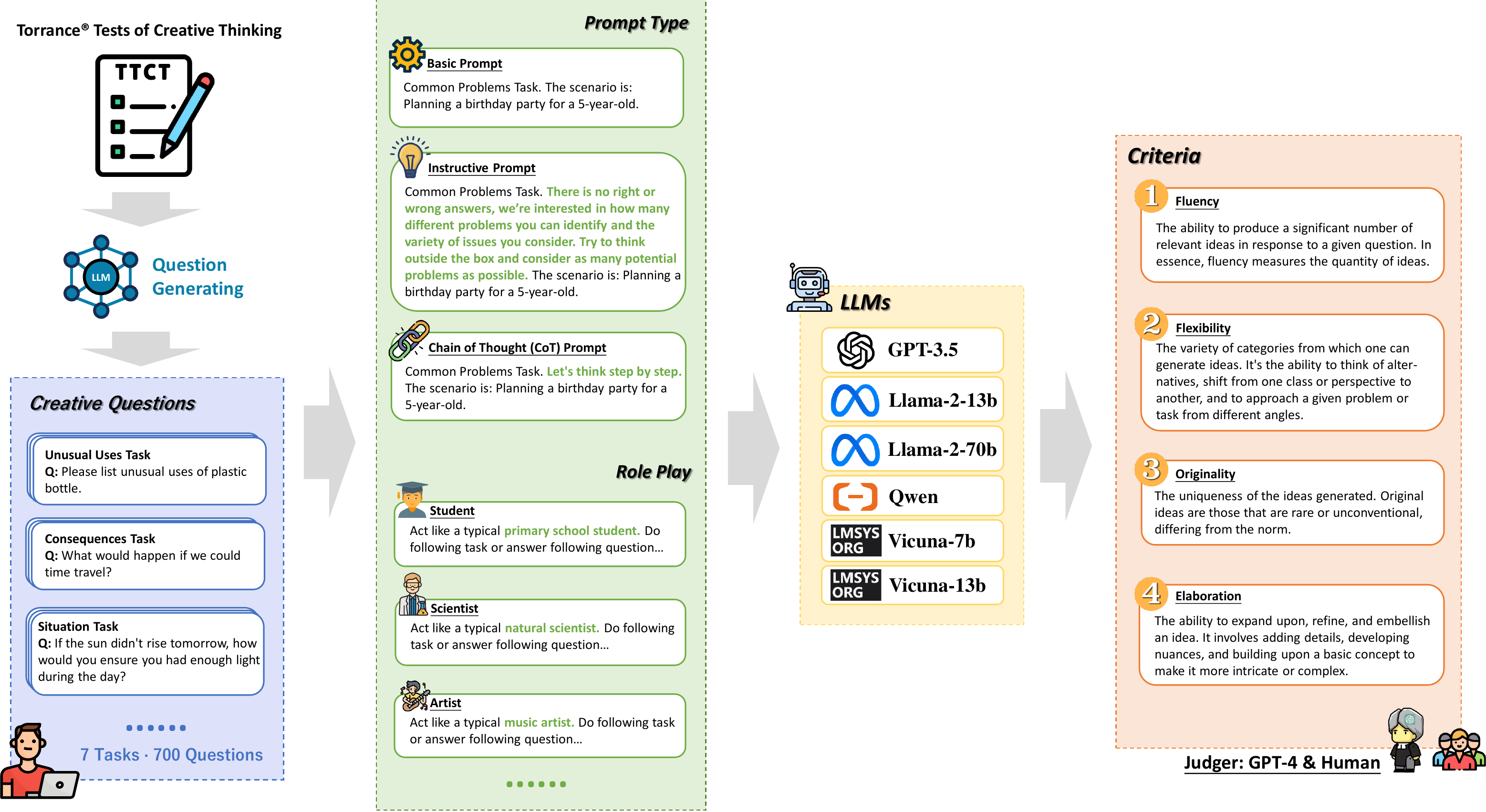}}
    \caption{This figure depicts the overall framework used to assess creativity in this paper. First, we referred to the task settings of the TTCT to generate our dataset, and then used this dataset to evaluate various large language models. The evaluation experiments included different prompts and comparative tests where LLMs played different roles. Finally, we used GPT-4 as an evaluator to assess the results of these models.}
    \label{fig:figframework}
\end{figure}
In this work, we design the overall framework to evaluate LLM's creativity, as shown in Fig. \ref{fig:figframework}.
Firstly, we construct a dataset containing 700 questions of 7 tasks  which was derived and modified from the psychology scale of the TTCT and expand the number of questions using GPT-4.
We tested six models on four different criteria using the dataset we constructed. Following this, we conducted a series of experiments on creativity of LLM when giving different types of prompts and assigning different roles to LLMs.
Finally, we used the GPT-4 as evaluator to gain the performance result of LLMs and verified the consistency of the LLM-based evaluation with humans.

\subsection{Dataset construction}\label{sec31}

This research utilized a modified TTCT verbal test, which are verbal tasks using verbal stimuli, includes seven distinct creative thinking tasks: (1) Unusual uses task; (2) Consequences task; (3) Just suppose task; (4) Situation task; (5) Common problem task; (6) Improvement task;  and (7) Imaginative stories task. The research suggests that there is cross-domain consistency in creativity and that there are common, creativity-related cognitive skills across domains. The seven tasks focusing on different aspects of creativity provide a good overall representation of an individual's level of creativity.

Each task includes 100 questions generated by GPT-4 using few-shot prompts. GPT-4 can generate a diverse and comprehensive set of similar problems based on the given examples, and all problems have been validated by humans to ensure usability. Detailed description of each task can be found in Method section.
\subsection{Evaluation criteria}\label{sec32}
We have four criteria for creativity evaluation:
\begin{itemize}
    \item \textbf{Fluency}. This refers to the ability to produce a significant number of relevant ideas in response to a given question. In essence, fluency measures the quantity of ideas.
    \item \textbf{Flexibility}. This assesses the variety of categories from which one can generate ideas. It's the ability to think of alternatives, shift from one class or perspective to another, and to approach a given problem or task from different angles.
    \item  \textbf{Originality}. This measures the uniqueness of the ideas generated. Original ideas are those that are rare or unconventional, differing from the norm.
    \item  \textbf{Elaboration}. This refers to the ability to expand upon, refine, and embellish an idea. It involves adding details, developing nuances, and building upon a basic concept to make it more intricate or complex.
\end{itemize}
Together, these criteria provide a comprehensive evaluation of an individual's creative abilities, considering not just the quantity of ideas they produce, but also the quality, diversity, and depth of those ideas.
\subsection{LLM-based evaluation}\label{sec33}
The standard TTCT evaluation method requires trained psychologists to follow professional manuals to assess the results, and an individual's single test only contains answers to a very limited number of questions. When evaluating creativity in LLM, both the insufficient sample of responses and the high human resource costs limit the application of creativity test on LLMs.

With the rapid development of LLM capabilities, the evaluation methods for many natural language processing tasks have evolved from traditional human annotation to reference-based automated methods, and now, to methods based on LLMs. LLMs are increasingly playing the role of judges in tasks such as question-answering, translation, and text quality assessment\cite{eval1,eval2}, giving rise to various evaluation framework\cite{eval3}. According to experimental results from relevant literature, LLM exhibits higher correlation with human evaluations compared to traditional automated technologies\cite{eval4,LLM_psyc}. In this study, based on the evaluation criteria from the previous section, we utilize GPT-4 to score the answer. For each criterion, the LLM needs to complete the Likert scale based on the responses. Additionally, we also verified the consistency between the evaluations made by LLM and human evaluations.

\section{Evaluation results}\label{sec4}
We conducted a statistical analysis of the creativity score of 6 popular LLMs across seven tasks, totaling 700 questions. We unveiled hidden conclusions within the data results from various dimensions. We compared the differences in creativity levels between the models, and we compared the performance variations under different criteria within the same model. 
Subsequently, we experimented with many types of prompts to see whether changes in prompts would affect the models' levels of creativity. Since LLMs possess the ability to play user-specified roles, we selected six typical human identities to explore the impact on creativity under different role-playing conditions. Finally, we utilized some psychological scales to test the LLMs, investigating the correlation between the personality traits of the LLMs and creativity.

\subsection{Results of different models and criteria}\label{sec41}

\begin{table}[]
\centering
\renewcommand\arraystretch{0.95}
\begin{tabular}{@{}ccccc@{}}
\toprule
                     & \textbf{Fluency} & \textbf{Flexibility} & \textbf{Originality} & \textbf{Elaboration} \\ \midrule
                     & \multicolumn{4}{c}{\textbf{Common Problem Task}}                                      \\ \midrule
\textbf{GPT-3.5}     & \textbf{4.975}   & \textbf{4.65}        & \textbf{3.87}        & 4.735                \\
\textbf{LLaMA-2-13b} & 4.94             & 4.48                 & 3.77                 & 4.89                 \\
\textbf{LLaMA-2-70b} & 4.92             & 4.545                & 3.72                 & \textbf{4.905}       \\
\textbf{Qwen}        & 3.09             & 2.89                 & 2.36                 & 3.36                 \\
\textbf{vicuna-13b}  & 4.91             & 4.32                 & 3.51                 & 4.415                \\
\textbf{vicuna-7b}   & 4.88             & 4.27                 & 3.38                 & 4.2                  \\ \midrule
\multicolumn{1}{l}{} & \multicolumn{4}{c}{\textbf{Consequences Task}}                                        \\ \midrule
\textbf{GPT-3.5}     & 4.855            & 4.81                 & \textbf{4.105}       & \textbf{5}           \\
\textbf{LLaMA-2-13b} & 4.91             & 4.83                 & 4.08                 & 5                    \\
\textbf{LLaMA-2-70b} & \textbf{4.93}    & \textbf{4.83}        & 3.995                & 4.995                \\
\textbf{Qwen}        & 4.41             & 4.43                 & 3.61                 & 4.875                \\
\textbf{vicuna-13b}  & 4.26             & 4.295                & 3.58                 & 4.85                 \\
\textbf{vicuna-7b}   & 4.535            & 4.435                & 3.66                 & 4.92                 \\ \midrule
\multicolumn{1}{l}{} & \multicolumn{4}{c}{\textbf{Improvement Task}}                                         \\ \midrule
\textbf{GPT-3.5}     & \textbf{5}       & \textbf{4.97}        & \textbf{4.62}        & \textbf{4.98}        \\
\textbf{LLaMA-2-13b} & 4.98             & 4.85                 & 4.15                 & 4.89                 \\
\textbf{LLaMA-2-70b} & 4.965            & 4.8                  & 4.085                & 4.9                  \\
\textbf{Qwen}        & 4.87             & 4.55                 & 3.76                 & 4.7                  \\
\textbf{vicuna-13b}  & 4.97             & 4.41                 & 3.6                  & 4.38                 \\
\textbf{vicuna-7b}   & 4.95             & 4.56                 & 3.86                 & 4.64                 \\ \midrule
\multicolumn{1}{l}{} & \multicolumn{4}{c}{\textbf{Imaginative Stories Task}}                                 \\ \midrule
\textbf{GPT-3.5}     & \textbf{4.16}    & \textbf{4.2}         & \textbf{4.475}       & \textbf{4.925}       \\
\textbf{LLaMA-2-13b} & 3.72             & 3.62                 & 4.03                 & 4.73                 \\
\textbf{LLaMA-2-70b} & 3.83             & 3.66                 & 4.05                 & 4.7                  \\
\textbf{Qwen}        & 3.24             & 3.51                 & 3.74                 & 4.43                 \\
\textbf{vicuna-13b}  & 3.31             & 3.61                 & 3.75                 & 4.49                 \\
\textbf{vicuna-7b}   & 3.28             & 3.47                 & 3.76                 & 4.58                 \\ \midrule
\multicolumn{1}{l}{} & \multicolumn{4}{c}{\textbf{Just Suppose Task}}                                        \\ \midrule
\textbf{GPT-3.5}     & \textbf{3.96}    & \textbf{4.31}        & 4.03                 & \textbf{4.93}        \\
\textbf{LLaMA-2-13b} & 3.83             & 4.16                 & \textbf{4.04}        & 4.93                 \\
\textbf{LLaMA-2-70b} & 3.795            & 4.09                 & 3.75                 & 4.87                 \\
\textbf{Qwen}        & 3.58             & 3.84                 & 3.25                 & 4.64                 \\
\textbf{vicuna-13b}  & 3.41             & 3.58                 & 3.03                 & 4.55                 \\
\textbf{vicuna-7b}   & 3.48             & 3.86                 & 3.24                 & 4.6                  \\ \midrule
\multicolumn{1}{l}{} & \multicolumn{4}{c}{\textbf{Situation Task}}                                           \\ \midrule
\textbf{GPT-3.5}     & 4.79             & 4.67                 & 3.94                 & 4.97                 \\
\textbf{LLaMA-2-13b} & 4.195            & 4.39                 & 3.92                 & 4.85                 \\
\textbf{LLaMA-2-70b} & \textbf{4.85}    & \textbf{4.8}         & \textbf{4.05}        & \textbf{4.99}        \\
\textbf{Qwen}        & 3.94             & 4.01                 & 3.17                 & 4.59                 \\
\textbf{vicuna-13b}  & 3.97             & 3.97                 & 3.14                 & 4.6                  \\
\textbf{vicuna-7b}   & 4.02             & 3.98                 & 3.21                 & 4.62                 \\ \midrule
\multicolumn{1}{l}{} & \multicolumn{4}{c}{\textbf{Unusual Uses Task}}                                        \\ \midrule
\textbf{GPT-3.5}     & \textbf{5}       & \textbf{4.92}        & \textbf{4.67}        & \textbf{4.895}       \\
\textbf{LLaMA-2-13b} & 4.99             & 4.86                 & 4.28                 & 4.91                 \\
\textbf{LLaMA-2-70b} & 4.98             & 4.85                 & 4.255                & 4.88                 \\
\textbf{Qwen}        & 4.905            & 4.21                 & 3.69                 & 4.13                 \\
\textbf{vicuna-13b}  & 4.86             & 4.06                 & 3.67                 & 3.76                 \\
\textbf{vicuna-7b}   & 4.91             & 4.64                 & 3.94                 & 4.3                  \\ \bottomrule
\end{tabular}
\caption{This table gives the average of all metrics across all tasks for all models. For each metric in each task, the values of the models with the highest scores are bolded.}
\label{table:overall table}
\end{table}

\begin{figure}[htbp]
    \centering
    \makebox[\textwidth][c]{\includegraphics[width=1.1\textwidth]{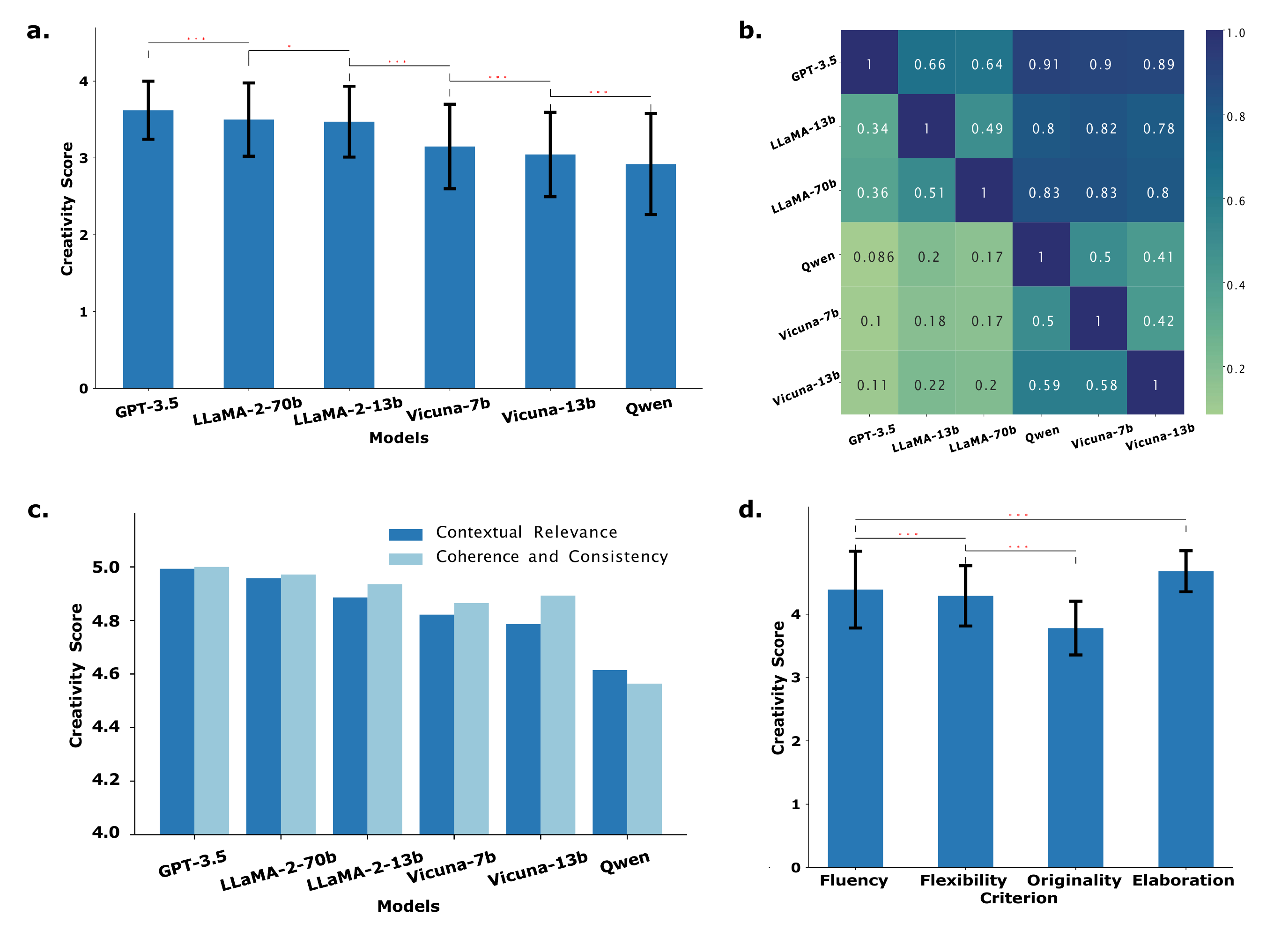}}
    \caption{\textbf{a.} Results of overall creativity scores of the seven models. The error bars represents the standard deviation. The measure of the centre for the error bars represents the average rating. Significance markers are placed above the bars, where \textbf{***} indicates a $p-value < 0.0001$, and \textbf{*} represents $p-value < 0.05$. Since the scoring data does not follow a normal distribution and is paired, the hypothesis test employed is the Wilcoxon signed-rank test. \textbf{b.} This is the heatmap of the win rate relationship between tested LLMs. The values in each grid represent the win rate of the model on the corresponding vertical axis compared to the model on the horizontal axis. \textbf{c.} This figure displays the scores for contextual relevance and coherence and consistency of models' answers. \textbf{d. }Results of overall creativity scores under four criteria. The error bars represents the standard deviation. The measure of the centre for the error bars represents the average rating. The statistical test is the same as the test in \textbf{a}.}
    \label{fig:h1}
\end{figure}

We had six models answer 700 questions and used GPT-4 evaluator to get scores for each model on all answers across various criteria. 
We first evaluate the average scores of each model across all tasks, as shown in Fig. \ref{fig:h1}a. It can be observed that GPT-3.5 exhibits the highest level of creativity, followed by the LLaMA-2 architecture models, then the LLaMA-based fine-tuned model vicuna, and finally qwen. The experimental results from the perspective of the model suggest that the type of the model have a significant impact on creativity, whereas the scale of parameters does not show a decisive influence. Different type of model means having different architecture and alignment strategy and using different datasets during the training process. These factors are likely to be key determinants of the level of creativity. 
Similar findings can also be observed in other LLM evaluation papers\cite{truthful,caremi,toolbench}. For example, in Toolbench\cite{toolbench}, the 30B version of LLaMA outperforms the 65B version of LLaMA in many tasks, and text-daVinci-003 also performs better overall than GPT-3.5.

To further validate the rank of the models, we conducted pairwise comparisons between the models, as shown in Fig. \ref{fig:h1}b. Each cell in this heatmap represents the win rate of the model on the y-axis in terms of creativity score compared to the model on the x-axis. The win rate scores are consistent with the strengths and weaknesses of the models shown in Fig. \ref{fig:h1}a, and we conducted statistical tests for significance, which are marked in the figure.

Next, we evaluate the average scores of each criterion across all tasks, as shown in Fig. \ref{fig:h1}d. The score for Elaboration is consistently high across all tasks, while Originality is relatively lower, with Fluency and Flexibility scoring in the middle. 
The capabilities of LLMs inherently stem from training on human language corpora, so it is intuitive that they score relatively lower in originality. The creativity of LLM is likely to be a manifestation of the combination of existing human knowledge, and how to improve the originality of LLM is an important future endeavor.
The Elaboration metric reflects the degree of refinement of a creative idea, and LLM's ability to articulate this has always been outstanding.

\begin{figure}[htbp]
    \centering
    \makebox[\textwidth][c]{\includegraphics[width=1.0\textwidth]{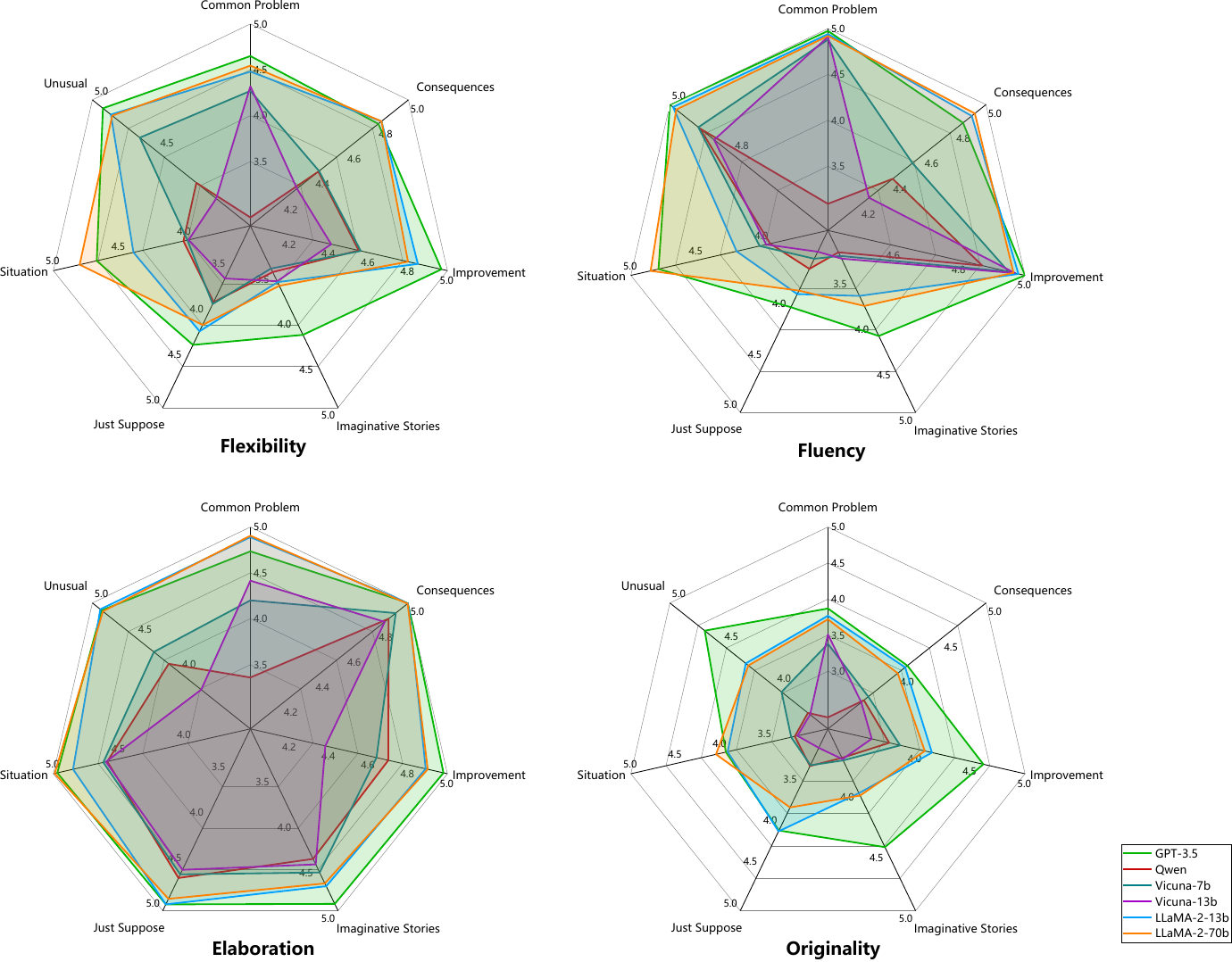}}
    \caption{The figure shows a radar chart of the performance of six models under four creativity assessment criteria across seven tasks. }
    \label{fig:radar}
\end{figure}

However, if we focus on the performance of different tasks, we will find that there are significant variances in creativity performance under different tasks, as shown in Fig. \ref{fig:radar}, which shows the radar charts of the performance of six models across seven tasks. It can be observed that most models exhibit a higher level of overall creativity in the Common Problem, Consequences, and Unusual Uses tasks, while the overall creativity level is lower in the Just Suppose and Imaginative Stories tasks, reflecting the varying degrees of creative difficulty presented by different tasks.

At last, there are some differences between humans and LLMs when answering questions. In the case of LLMs responding to human prompts, issues such as irrelevance to the topic or logical errors may arise. On the other hand, humans generally maintain consistency in their answers. So we have evaluated the responses of all models in this regard, and it is observable that there are significant differences in relevance and coherence among the various models, as shown in Fig. \ref{fig:h1}c. The results show that the GPT-3.5 and LLaMA models performed well, while the Vicuna and Qwen models had poorer performance. Sometimes, Vicuna and Qwen fail to understand the question properly, leading to irrelevant answers. Sometimes, due to a misunderstanding of the question, they refuse to answer. We all consider these as manifestations of a lack of creativity. This issue may be related to the alignment strategies employed and the training datasets of the models.

\subsection{Results of different prompt types}\label{sec42}
The prompt is a crucial component of the LLM model, as it provides the necessary context and information for LLMs to generate a relevant and coherent response. The quality and type of the prompt can significantly impact the quality of the generated response. Therefore, we believe that the type of prompt can greatly influence the creativity of LLMs.

In our experiment, we designed and compared four different types of prompts: basic prompt, instructive prompt, post-instructive prompt and Chain of Thought (CoT) prompt. Herein, the basic prompt contains only the essential information needed to describe the task, simple and clear. The instructive prompt provides a detailed description of the expected answer, outlining what constitutes a creative response. The post instruction prompt uses two rounds of prompts, starting with a basic prompt for the LLM to give a basic answer, then giving some instruction about creativity (the same as instructive prompt). LLM revises the answer given in the first round based on the given instruction and gives the revised answer. Chain of Thought is a technique that enables complex reasoning capabilities through intermediate reasoning steps or just a single explicitly prompt like "Let's think step by step". We utilize the technique used in \cite{COT} to design our CoT prompt. The example of prompts are shown in Fig. \ref{fig:figframework}.

\begin{figure}[htbp]
    \centering
    \makebox[\textwidth][c]{\includegraphics[width=0.9\textwidth]{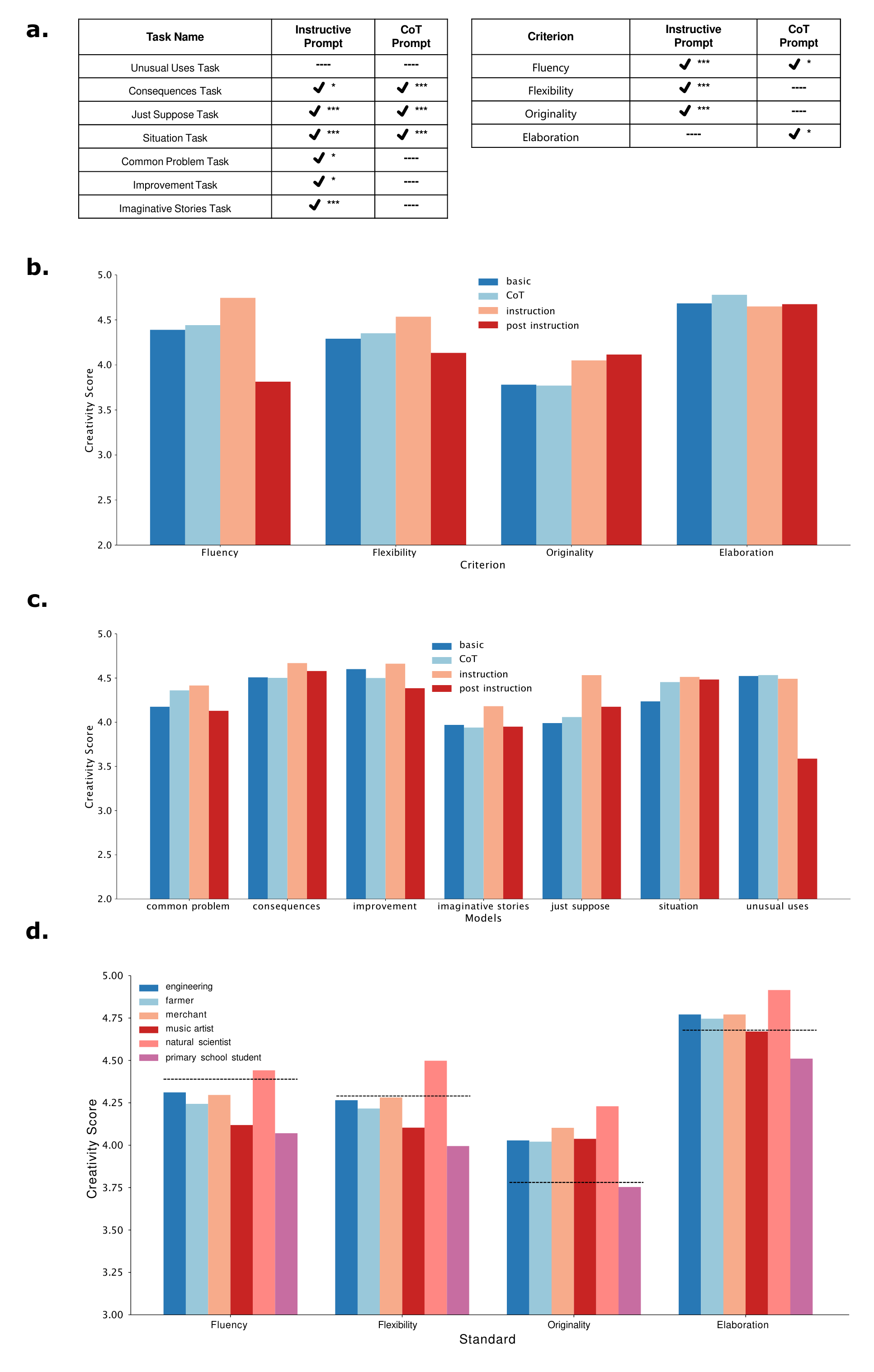}}
    \caption{\textbf{a.} In this figure, we have compiled statistics on the impact of different prompt types across various tasks and according to different criterion of creativity. Herein, a '$\checkmark$' signifies an enhancement in creativity, while a '---' indicates no significant effect. Significance markers are placed near '$\checkmark$', where \textbf{***} indicates a $p-value < 0.0001$, and \textbf{*} represents $p-value < 0.05$ which is calculated by Wilcoxon signed-rank test. \textbf{b.} This figure depicts the performance of creativity across all criteria for different types of prompts. All the hypothesis tests have been given in the figure above. \textbf{c. } This figure depicts the performance of creativity across all tasks for different types of prompts. All the hypothesis tests have been given in the figure above. \textbf{d.} The figure illustrates the values for each creativity metric of the LLM across all tasks when assigned different roles. The horizontal line in the figure indicates the level of creativity of the LLM without any role-play system prompt.}
    \label{fig:h2}
\end{figure}

As shown in Fig \ref{fig:h2}a, b and c, We have obtained data on the performance of LLMs in terms of creativity across all tasks and all criteria under different prompt types. From the perspective of the task, the inclusion of instructive language in prompts has improved creativity in all tasks except for `unusual uses'. The reason for no improvement in `unusual uses' may be that the task description is already clear enough and the required divergent thinking ability is relatively simple. When using the CoT (Chain of Thought) prompt, there has been an increase in the level of creativity in three tasks, indicating that some tasks require a higher level of convergent thinking ability to demonstrate creativity. In the case of post instruction, the greatest differences were shown between tasks. While a few tasks, such as imaginative stories and just suppose, showed some rise, most of the rest did not have a significant boosting effect, and even produced a drastic drop on the unusual uses task. We speculate that the main reason may be that under multiple rounds of dialog, the post-instruction actually implicitly negates the initial response, resulting in the inability to be in a position to come up with a more creative response on a relatively simple task, such as unusual uses.
From the perspective of creativity criteria, it is evident that instructive prompts significantly enhance both flexibility and originality but do not increase elaboration. On the other hand, CoT prompts slightly improve elaboration. Both types of prompts are beneficial to the fluency of the responses. In summary, the creative performance of LLMs, like their other abilities, is significantly influenced by the prompts. Effective prompt engineering is greatly beneficial for better harnessing the potential creativity of LLMs.
For the post instruction prompt, only originality criteria has obvious increase, and even a significant decrease in the fluency and flexibility criteria. For the same reason as previously stated, the second round of responses will naturally negate the initial responses, resulting in a lack of flexibility and fluency in the final answer.

\subsection{Results of playing different roles}\label{sec43}
LLMs possess the remarkable capability to adopt the roles specified by users, which can subsequently influence their outputs. This adaptability enables the models to deliver tailored responses, aligning with the context and characteristics of the assumed identities. In our experiment, we attempted to specify the exact identity and role of the LLM within the system prompt. The primary objective of this approach was to ascertain whether the LLM could enhance its creative expression by adopting specific roles and to determine if this influence is consistent with cognitive patterns observed in reality.

As shown in \ref{fig:h2}d, We assigned six distinct roles to the LLM: an engineering, a farmer, a merchant, a scientist, an artist, and an primary school student, requiring the model to perform tasks in alignment with the characteristics of the respective roles. The results demonstrated that across all creativity assessment criteria, the creativity level of the scientist surpassed that of the other six roles, reflecting a correlation between the accumulation of knowledge, educational attainment, and the level of creativity. Furthermore, when LLM was playing different roles except scientist, the values of fluency and flexibility have decreased, yet originality has increased significantly. This suggests that giving LLM specific roles induces more original responses. This experiment reveals the weakness of LLM's lack of originality in its default situation.

\subsection{Results of creativity under collaboration}
\begin{figure}[htbp]
    \centering
    \makebox[\textwidth][c]{\includegraphics[width=0.9\textwidth]{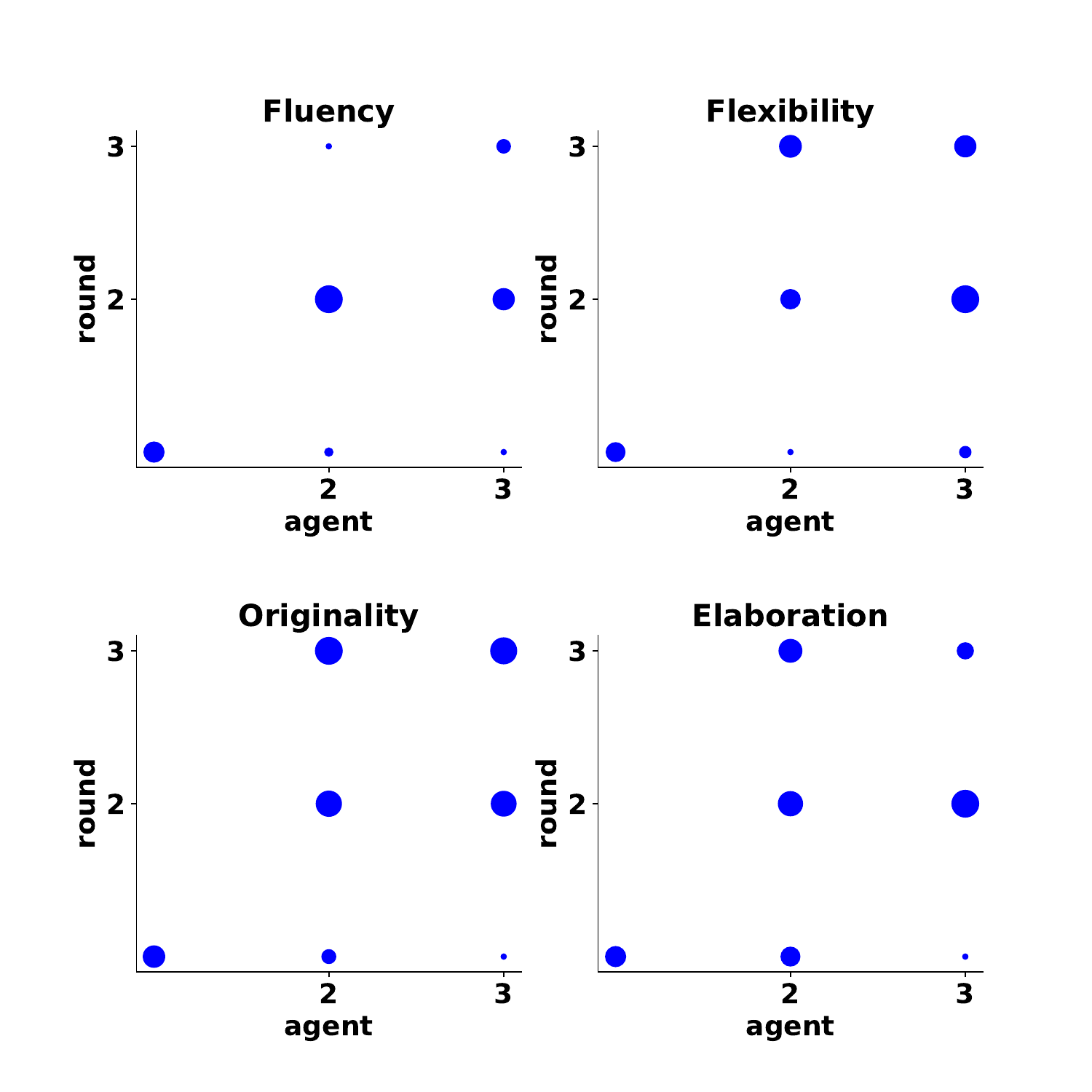}}
    \caption{In this figure, we presented scatter plots of the creativity scores under different criteria, varying by the number of rounds and agents. The area of each scatter point represents the level of creativity.}
    \label{fig:collab}
\end{figure}
In reality, creative activities can be accomplished through collaboration and discussion among multiple individuals. Literature indicate that the process of creative collaboration can enhance and increase the innovativeness of the outcomes\cite{collab1,collab2}. Inspired by this, we believe that results produced through the collaboration of LLMs have stronger creativity compared to those generated by a single LLM.

Based on the above analysis, in this section, we attempt to have multiple agents engage in multi-round discussions about questions in the dataset, ultimately providing a joint final answer. After the previous LLM provides an answer, the subsequent LLM will use that answer as inspiration to give its own response. Once a predetermined number of rounds is reached, the final result is presented.
In our experiment, using GPT-3.5 as the base model, we explored the changes in scores under different creativity criteria when the number of LLMs is 2 and 3 (we call it agent) and the number of rounds is 2 and 3. We compared these scores with the creativity scores obtained under default conditions.

As shown in Fig. \ref{fig:collab}, We presented scatter plots of the creativity scores under different criteria, varying by the number of rounds and agents. The area of each scatter point represents the level of creativity. From the results, we can see some interesting findings: First, when the number of rounds is one, an increase in the number of agents leads to a decrease in the level of creativity across the four criteria. This might be due to the lack of multiple reviews of the answers by the same agent in a single round of interaction, leading to the answers of later-ranked agents constantly negating previous answers. Additionally, in the cases of originality, flexibility, and elaboration, an increase in both rounds and agents enhances the level of creativity, with the most significant improvement observed in originality. This supports the conclusion that collaboration can enhance creativity and is consistent with human behavior. Lastly, there are some exceptions to the above conclusion, such as a decrease in fluency when there are two agents and three rounds. This could be due to excessive discussion making the answers overly concise.

\subsection{Investigation of the relationship between LLM's creativity and its personality traits}\label{sec44}

\begin{table}[]
\resizebox{\linewidth}{!}{
\begin{tabular}{@{}cccccc@{}}
\toprule
                          &                         & Emotional Intelligence & Empathy           & Self-Efficacy     & Openess           \\ \midrule
\multirow{2}{*}{Kendall $\tau$}  & Correlation Coefficient & \textbf{0.3825}        & \textbf{0.3825}   & \textbf{0.4401}   & 0.0329            \\
                          & p-value                 & \textless{}0.0001      & \textless{}0.0001 & \textless{}0.0001 & 0.6054            \\
\multirow{2}{*}{Spearman $\rho$} & Correlation Coefficient & \textbf{0.5026}        & \textbf{0.5026}   & \textbf{0.5613}   & 0.0526            \\
                          & p-value                 & \textless{}0.0001      & \textless{}0.0001 & \textless{}0.0001 & 0.5369            \\ \midrule
                          &                         & Conscientiousness      & Extraversion      & Agreeableness     & Neuroticism       \\
\multirow{2}{*}{Kendall $\tau$}  & Correlation Coefficient & \textbf{0.3691} & \textbf{0.2636} & \textbf{-0.37}   & \textbf{0.2533} \\
                          & p-value                 & \textless{}0.0001      & \textless{}0.0001 & \textless{}0.0001 & \textless{}0.0001 \\
\multirow{2}{*}{Spearman $\rho$} & Correlation Coefficient & \textbf{0.4897} & \textbf{0.3394} & \textbf{-0.4967} & \textbf{0.3455} \\
                          & p-value                 & \textless{}0.0001      & \textless{}0.0001 & \textless{}0.0001 & \textless{}0.0001 \\ \bottomrule
\end{tabular}
}
\caption{This figure shows the correlation of our selected personality traits with their level of creativity on LLMs. We report the coefficients for Kendall $\tau$ and Spearman $\rho$ and p-values to ensure significance. Bold font indicates that the item is significant.}
\label{table:psych}
\end{table}
We have mentioned that, creativity evolves from a combination of individual processes such as cognitive, affective, behavioral, and contextual factors. In this section, we subject LLMs to a series of psychometric tests traditionally used to assess human personality traits. Our aim is to explore whether, akin to humans, there is a correlation between various personality factors and the creative capabilities of these advanced computational systems.

In our experiment, we have selected eight personality traits: emotional intelligence, empathy, self-efficacy, openness, conscientiousness, extraversion, agreeableness, and neuroticism. The latter five are the classic Big Five personality traits. Meta-analytic literature\cite{meta1,meta2} in the field of psychology suggests that each of these eight traits correlates with levels of creativity.

As shown in Table. \ref{table:psych}, we conducted experiments on LLMs and reported the correlations between the mentioned personality traits and creativity. We chose the Kendall $\tau$ and Spearman $\rho$ as the correlation coefficients and performed hypothesis testing. The experimental results indicate that the levels of emotional intelligence, empathy, conscientiousness, extraversion, and neuroticism in Large Language Models have a significant positive correlation with creativity levels, while agreeableness shows a significant negative correlation. Apart from agreeableness and openness, the influence of the remaining personality traits on creativity in large models is consistent with human performance.
\section{Discussion}\label{sec5}
In this article, we have presented a framework to assess and understand the creativity of LLMs. The core of this framework consists of 7 tasks and LLM-based evaluation protocol that can be used to assess LLM's creativity along four criteria. The proposed framework can be used to assess the creative performance of LLMs from multiple dimensions, while also exploring the factors that influence the creativity of these models and the relationships with other model characteristics. To illustrate the use and usefulness of our framework, we constructed a dataset containing 700 questions that encompass various types of tasks measuring divergent thinking. 

Through our further analysis and experiments, we demonstrated that the creativity of LLMs is significantly influenced by the type of model architecture, the type of prompts it receives, and the model's system prompts. At the same time, we also revealed a correlation between the levels of creativity of LLMs and their personality traits. This work is beneficial for our deeper understanding of the representations of LLMs and try to establish a bridge between artificial intelligence models and human cognitive models.

Although we have proposed an effective framework for measuring the creativity of LLMs, it still has some limitations that need to be addressed by future work.
Firstly, LLMs use text as both input and output, which allows them to borrow from psychological methods of creativity assessment such as TTCT, which using verbal task with verbal stimuli. However, with the rapid development of AI models, those accepting multimodal inputs are emerging\cite{multimodal}, which we call large multi-modal model (LMM). Designing a variety of tasks beyond verbal question-and-answer formats for assessing the creativity of these LMM is an important direction for future research. 
Secondly, LLMs are not the only generative models capable of expressing creativity; there are also image generation models based on diffusion models and models for generating music\cite{diffusion,music_generation}, in another words, can generate multi-modal outputs. How to assess the content produced by these other types of models to measure their level of creativity is also a question worth considering. 
In addition, The power of LLM allows developers to use it to develop a wide variety of plug-ins, integrate it with external programs or software, and even construct an agent system \cite{agent_survey}, and the creativity in these cases is bound to be different and needs to be investigated.
Lastly, we believe that the creativity exhibited by LLMs is only an outcome-oriented interpretation. Whether AI models possess true creativity from a human cognitive perspective remains an open question in the field of artificial intelligence. LLM's expression of creativity is likely to be an imitation of human creativity through a large number of learning.
Understanding the creativity of LLMs is also beneficial for uncovering the inner secrets of the model `black box', and for a deeper human understanding of the nature of intelligence and cognition. 
Although analyzing the nature of creativity is difficult, our analysis and evaluation of LLM creativity performance is fundamental to the study of the kernel of creativity.

\section{Methods}\label{sec6}

\subsection{Tested Models}
We tested six of the most advanced LLMs and listed as below. All the models were implemented with the open-source repository HuggingFace\cite{huggingface}.
\begin{itemize}
    \item \textbf{GPT-3.5. } GPT-3.5 is a language model developed by OpenAI, which is an advanced version of the GPT-3 model. It is capable of generating natural language text and code. GPT-3.5 were trained on an Azure AI supercomputing infrastructure. The version we used in the experiments are GPT-3.5-turbo-0613.
    \item \textbf{LLaMA-2. } LLaMA-2 is a family of state-of-the-art open-access large language models released by Meta and Microsoft\cite{llama2}. It is built upon the success of its predecessor, LLaMA-1. LLaMA-2 is specifically designed to facilitate the development of generative AI-powered tools and experiences. It is available for free for research and commercial use. The Llama 2 release introduces a family of pretrained and fine-tuned LLMs, ranging in scale from 7B to 70B parameters. The versions we used in the experiments are LLaMA-2-13b-chat-hf and LLaMA-2-70b-chat-hf.
    \item \textbf{Vicuna. } Vicuna is a lightweight, accurate, and efficient language model developed by a team of researchers from several universities, including UC Berkeley, Carnegie Mellon University, Stanford, and UC San Diego\cite{eval2}. It was built from Meta’s adaptable LLaMA model, which was fine-tuned on a dataset of around 70,000 human-generated conversations from the ShareGPT website. The versions we used in the experiments are Vicuna-7b-v1.5 and Vicuna-13b-v1.5.
    \item \textbf{Qwen. } Qwen (abbr. Tongyi Qianwen), proposed by Alibaba Cloud\cite{qwen}. It is a Transformer-based large language model, which is pretrained on a large volume of data, including web texts, books, codes, etc. The versions we used in the experiments are Qwen-7b-chat.
\end{itemize}
\subsection{Tasks using for creativity assessment}
This research utilized a modified TTCT verbal test, which are verbal tasks using verbal stimuli, includes seven distinct creative thinking tasks: (1) Unusual uses task; (2) Consequences task; (3) Just suppose task; (4) Situation task; (5) Common problem task; (6) Improvement task;  and (7) Imaginative stories task. Each task includes one hundred questions generated by GPT-4 using few-shot prompts. The seven tasks were generally structured as follows:
\begin{itemize}
    \item \textbf{Task 1: Unusual uses. }This task challenges individuals to think of as many unusual and diverse uses as possible for a common object within a limited time frame. The object in question is typically everyday and familiar, such as a brick, paperclip, or newspaper.
    \item \textbf{Task 2: Consequences. }This task focuses on the ability to foresee consequences or outcomes of an unusual or hypothetical situation. For example, What would be the implications if animals could talk?
    \item \textbf{Task 3: Just suppose. }This task encourages imaginative and speculative thinking by asking participants to consider hypothetical, often fantastical, scenarios and their implications. For example, just suppose you woke up one morning and found you could fly. What would you do? List as many things as you can think of.
    \item \textbf{Task 4: Situation task. }This task is designed to assess creative thinking by evaluating how individuals respond to and interpret a given situation. This task emphasizes understanding social dynamics, empathy, and the ability to think of multiple perspectives or solutions. For example, if all books were to disappear, how would you gain knowledge?
    \item \textbf{Task 5: Common problem. }This task focuses on everyday problems that are familiar to most people, requiring participants to generate innovative and effective solutions. For example, Organizing a cross-country road trip or building a tree house.
    \item \textbf{Task 6: Improvement. }This task focused on assessing an individual's ability to enhance or modify existing objects or ideas. The given object is similar with the unusual uses task.
    \item \textbf{Task 7: Imaginative stories. }This task is designed to assess creativity through narrative and storytelling with a given prompt. This task emphasizes the ability to construct original, coherent, and imaginative stories, showcasing an individual's creative potential in terms of narrative ability. Examples of given prompts are "The Invisible Elephant" or "The Book that Wrote Itself".
\end{itemize}
\subsection{Detail of experiment settings}
The models used in our experiment primarily originate from the open-source HuggingFace platform. The specific versions of these models have already been reported above. In this section, we will specifically present the experimental parameters and other settings related to the experiment.

For a LLM based on the Transformer architecture, there are certain parameters that directly affect the output of the model. 
\begin{itemize}
    \item \textbf{max tokens. }This parameter controls the maximum number of tokens to generate in the chat completion. In our experiment, this value is uniformly set to 512, ensuring that the output length is sufficient to maintain the quality of the answers.
    \item \textbf{temperature. }The parameter is a crucial factor in determining the nature of the model's responses. This is a hyperparameter that influences the randomness or unpredictability in the model's responses. Essentially, its mechanism is to change the probability distribution of the model's output logits. However, according to our experiments, changes in temperature do not significantly affect creative performance, which appears quite random. Therefore, in our experiments, the temperature is uniformly set to 1.
    \item \textbf{top\_p. }Top-p is also a parameter used to control the diversity of generated text, also known as 'nucleus sampling'. This parameter's full name is 'top probability', which is typically represented by a value between 0 and 1, indicating the cumulative threshold of the highest probabilities chosen in the probability distribution when generating the next token. In our experiments, top\_p is uniformly set to 1.
    \item \textbf{top\_k. }This parameter is used when generating the next token to limit the model to only consider the top k tokens with the highest probability. This strategy can reduce the likelihood of the model generating meaningless or repetitive outputs, while also improving the speed and efficiency of the model's generation. In our experiments, top\_k is uniformly set to 50.
\end{itemize}

GPT-4 serves as the judge for our LLM-based evaluation, with its relevant parameters set to default. The version used is GPT-4-0613. Besides, all prompt templates used in the experiment will be provided in the appendix.
\subsection{Psychology scales}
In this experiment, we explored the relationship between personality traits and creative performance of some large models, using some public psychological scales and related literature. 

We use the Situational Test of Emotion Management (STEM) for the assessment of emotional intelligence\cite{EI}. STEM evaluates an individual's ability to manage emotions in various situations. It's based on the concept of emotional intelligence, which involves recognizing, understanding, and managing one's own emotions and those of others. The test typically presents a series of hypothetical scenarios to the participants. Each scenario is designed to assess different aspects of emotional intelligence, such as emotional awareness and regulation. Participants are asked how they would respond to each situation. Their responses are then analyzed to determine their EI levels. STEM is used in various settings, including organizational training, psychological research, and personal development. It helps in identifying areas where emotional intelligence can be improved, which is valuable in both personal and professional contexts.

We use Toronto Empathy Questionnaire (TEQ) for assessing LLM's empathy level\cite{empathy}. The TEQ was developed by researchers at the University of Toronto. It's grounded in the idea that empathy is a multi-dimensional construct, involving both cognitive and affective elements. The questionnaire consists of 16 items, each rated on a 5-point Likert scale. These items are designed to measure the respondent's emotional and cognitive responses to the experiences and feelings of others. The TEQ is used in various fields, including psychological research, clinical settings, and social science studies. It helps in understanding how individuals emotionally connect with others, which can be important in contexts like therapy, counseling, and social work.

We use Generalized Self-Efficacy scale\cite{GSE} to assess LLM's self-efficacy. The scale was developed by Ralf Schwarzer and Matthias Jerusalem in 1995. It's part of a larger body of research on self-efficacy and psychological well-being. The Generalized Self-Efficacy Scale is a short survey consisting of 10 items. Respondents rate each item on a scale, typically from 1 to 4, where higher scores indicate greater self-efficacy. Unlike scales that measure task- or situation-specific self-efficacy, this scale assesses a general sense of personal competence to deal effectively with a variety of stressful situations. It is widely used in psychological research, clinical psychology, and health psychology. It's also utilized in organizational and educational settings to understand and enhance individuals' beliefs in their own capabilities. The Generalized Self-Efficacy Scale has been validated in numerous studies across different cultures and is known for its reliability and construct validity.

Finally, we applied the classic Big Five Inventory test to LLMs\cite{BFI}. The Big Five personality traits, include Openness, Conscientiousness, Extraversion, Agreeableness, and Neuroticism (often abbreviated as OCEAN). These traits represent a broad range of human personality characteristics and are believed to be universal. The BFI typically consists of short phrases or statements that respondents rate on a scale in terms of how well they describe their own behavior or characteristics. The BFI is valued for its balance between brevity and comprehensive coverage of the five-factor model. It demonstrates good reliability and validity, making it a trusted tool in personality assessment.
\subsection{Model-human agreement evaluation}
To confirm that the assessment methods based on LLMs are overall reasonable and consistent with human judgement, we sample the responses generated by these models and hire humans to evaluate them. We presented the answer pairs generated by the LLMs to 20 native English-speaking participants globally (10 male) recruited from Prolific (\href{https://www.prolific.co/}{www.prolific.co}), and paid each participants \pounds 15. The average reward per hour for the participants is \pounds 14.59. The average participants age was 32.9 $\pm$ 20.1. In the experiment, we sampled seven tasks, resulting in 84 pairs of questions and answers, which means there are 84 trails. These pairs consist of answers from different models to the same question within the same task, and are presented to the participants.

On each trail of the task, participants were asked to make a binary decision about which of the two answers is more creativity according to the given criteria. Participants also have the option to choose that there is no significant difference in creativity between the two responses. A progress bar at the top of the screen indicated to participants how many trails they had completed and had remaining to complete. After obtaining the final human evaluation data, we calculate the consistency between the human assessment results and those of the LLMs. We use Kendall's coefficient and Spearman's coefficient for this calculation. Since the participants' data is based on relative win-loss relationships, we need to preprocess the human evaluation results. For tie results, we convert the human assessment results to the average score two answers evaluated by LLM; for non-tie results, we assign the higher score evaluated by the LLM to the winning response in the human results. Under these conditions, the calculated Kendall's coefficient and Spearman's coefficient are 0.4996 and 0.5564, respectively. These values are quite usable for automated evaluation techniques.

\bibliographystyle{naturemag}
\bibliography{main}

\clearpage


\clearpage

\end{document}